\documentclass[conference]{IEEEtran}
\IEEEoverridecommandlockouts

\usepackage{cite}
\usepackage{amsmath,amssymb,amsfonts}
\usepackage{algorithmic}
\usepackage{graphicx}
\usepackage{textcomp}
\usepackage{graphicx}
\usepackage{xcolor}

\usepackage[table]{xcolor} 
\usepackage{booktabs}      
\usepackage{multirow}      
\usepackage{xcolor}
\def\BibTeX{{\rm B\kern-.05em{\sc i\kern-.025em b}\kern-.08em
    T\kern-.1667em\lower.7ex\hbox{E}\kern-.125emX}}
\begin{document}

\title{Beyond Random Masking: A Dual-Stream Approach for Rotation-Invariant Point Cloud Masked Autoencoders

}

\author{\IEEEauthorblockN{Xuanhua Yin, Dingxin Zhang, Jianhui Yu, Weidong Cai}}
\author{Xuanhua Yin, Dingxin Zhang, Yu Feng, Shunqi Mao, Jianhui Yu, Weidong Cai\\
    School of Computer Science, The University of Sydney\\
    {\tt \{xuanhua.yin,dingxin.zhang,yfen0146,smao7434,jianhui.yu,tom.cai\}@sydney.edu.au} 
}

\maketitle

\begin{abstract}
Existing rotation-invariant point cloud masked autoencoders (MAE) rely on random masking strategies that overlook geometric structure and semantic coherence. Random masking treats patches independently, failing to capture spatial relationships consistent across orientations and overlooking semantic object parts that maintain identity regardless of rotation. We propose a dual-stream masking approach combining 3D Spatial Grid Masking and Progressive Semantic Masking to address these fundamental limitations. Grid masking creates structured patterns through coordinate sorting to capture geometric relationships that persist across different orientations, while semantic masking uses attention-driven clustering to discover semantically meaningful parts and maintain their coherence during masking. These complementary streams are orchestrated via curriculum learning with dynamic weighting, progressing from geometric understanding to semantic discovery. Designed as plug-and-play components, our strategies integrate into existing rotation-invariant frameworks without architectural changes, ensuring broad compatibility across different approaches. Comprehensive experiments on ModelNet40, ScanObjectNN, and OmniObject3D demonstrate consistent improvements across various rotation scenarios, showing substantial performance gains over the baseline rotation-invariant methods.
\end{abstract}


\section{Introduction}
Self-supervised learning, especially with Masked Autoencoders (MAE), has shown great promise in 3D point cloud representation learning~\cite{pang2022masked}. However, existing point cloud MAE methods face critical limitations under arbitrary rotations, which are common in real-world scenarios~\cite{feng2023masklrf,su2025ri}. Specifically, models trained on aligned point clouds experience significant performance degradation when tested on rotated versions of the same objects, failing to recognize identical shapes in different orientations.

\begin{figure}[t]
\centering
\includegraphics[width=\linewidth]{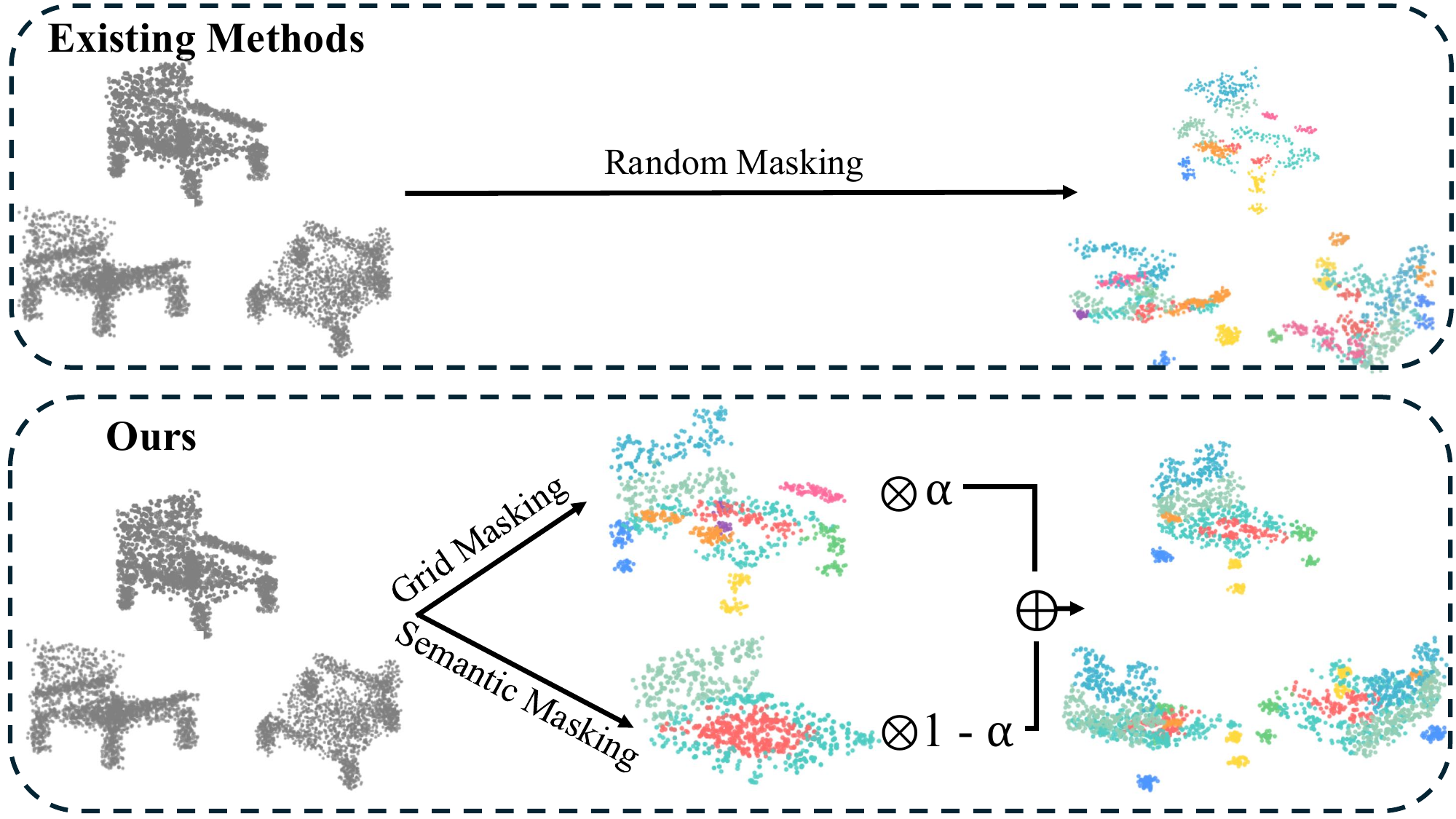}
\caption{Comparison between existing random masking and our dual-stream masking approach. \textbf{(Top)} Existing methods use random masking that treats patches independently, failing to preserve geometric structure and semantic coherence across different orientations. \textbf{(Bottom)} Our approach combines Grid Masking for spatial structure and Semantic Masking for part-based understanding through dynamic weighting ($\alpha$ and $1-\alpha$), producing consistent rotation-invariant masking patterns.}
\label{fig:intro}
\end{figure}

Models trained with standard MAE approaches suffer from significant performance drops under rotational variations, highlighting the need for specialized strategies that can handle arbitrary orientations effectively. To address this challenge, recent MAE methods extract rotation-invariant features and design relative position encodings~\cite{feng2023masklrf, su2025ri, yin2025hfbri,YuAAAI23_RethinkRI}. However, these approaches universally employ random masking strategies while overlooking the impact of masking methodology on enhancing rotation-invariant MAE model performance. 
Random masking encourages general robustness, but it may under-represent two aspects important for RI learning in practice: (i) preserving spatial/geometric relations and (ii) maintaining semantic coherence of object parts. 

We propose that effective rotation-invariant learning requires explicitly preserving these two aspects during the masking process. Our key insight is that the masking strategy itself must respect invariant properties rather than treating them as stochastic elements. To address the limitations of random masking, we propose a paradigm that systematically preserves the properties that should remain invariant under rotation. We propose to design masking strategies that inherently respect both the geometric structure and semantic coherence of point clouds. Rather than treating each patch in isolation, we consider the broader context of spatial relationships and semantic unity. This conceptual approach ensures that the masking process aligns with the fundamental requirements of rotation-invariant learning, enabling models to focus on truly invariant characteristics rather than orientation-dependent artifacts. By addressing the root cause of the mismatch between random masking and rotation-invariant objectives, our approach enhances the learning of robust representations that remain consistent across arbitrary orientations.

To realize this principled masking strategy, we introduce a dual-stream framework that operationalizes both spatial and semantic preservation requirements. As illustrated in Figure~\ref{fig:intro}, our approach combines 3D Spatial Grid Masking with Progressive Semantic Masking through curriculum learning~\cite{bengio2009curriculum}. The spatial grid component organizes patches according to their coordinate relationships, creating structured masking patterns that respect geometric consistency across rotations. The semantic component employs attention-driven clustering~\cite{vaswani2017attention} to identify and preserve functionally coherent object parts during masking. A dynamic weighting mechanism balances these two streams throughout training, beginning with emphasis on geometric structure and gradually incorporating semantic understanding as the model matures. The entire framework is designed as plug-and-play components that enhance existing rotation-invariant MAE architectures~\cite{feng2023masklrf, su2025ri, yin2025hfbri} without requiring structural modifications, ensuring broad compatibility and practical deployment across diverse rotation-invariant approaches. Random masking remains an effective and widely used pretext because of its unbiased sampling and unpredictability, often leading to robust representations. Our dual-stream masking, by contrast, introduces an explicit spatial--semantic prior tailored to rotation-invariant (RI) objectives. This prior can improve invariance under diverse orientations but also constitutes a bias that may trade off some task-agnostic generality. We therefore view our approach as a complementary alternative when rotation invariance is prioritized, rather than a universal replacement for random masking.

The main contributions of this work are:
\begin{itemize}
    \item We propose 3D Spatial Grid Masking and Progressive Semantic Masking to preserve geometric structure and object part coherence, addressing random masking limitations in rotation-invariant scenarios.
    \item We introduce a dual-stream curriculum learning framework that synergistically combines these strategies, progressing from geometric to semantic understanding for enhanced rotation-invariant representation learning.
    \item We demonstrate effectiveness across multiple rotation-invariant MAE frameworks, achieving consistent improvements on ModelNet40~\cite{wu20153d}, ScanObjectNN~\cite{uy2019revisiting}, and OmniObject3D~\cite{wu2023omniobject3d} datasets.
\end{itemize}

\section{Related Work}

\subsection{Self-supervised Learning for Point Clouds}
Self-supervised learning (SSL) provides a powerful paradigm for learning meaningful representations from unlabeled 3D point cloud data~\cite{zeng2024self}. Early contrastive methods establish frameworks through positive-negative pair learning. PointContrast~\cite{xie2020pointcontrast} and CrossPoint~\cite{afham2022crosspoint} represent typical examples. Recent advances adopt attention-driven approaches. PointSL~\cite{zhou2024pointsl} and PointACL~\cite{pointacl2024} demonstrate this trend. Masked autoencoding achieves remarkable success in point cloud learning. Point-MAE~\cite{pang2022masked} establishes the foundation and adapts random masking through FPS-based patch generation. Point-M2AE~\cite{zhang2022point} extends this approach with multi-scale hierarchical pretraining. Point-CMAE~\cite{ren2024bringing} integrates contrastive properties into the framework. While 2D vision methods like MAE~\cite{he2022masked} and BEiT~\cite{bao2021beit} employ various masking strategies, point clouds present unique challenges due to irregular structure and varying density. 


However, rotation invariance significantly impacts the effectiveness of point cloud MAE methods. MaskLRF~\cite{feng2023masklrf} uses Local Reference Frames but introduces computational overhead. RI-MAE~\cite{su2025ri} adopts PCA-based alignment but suffers from noise sensitivity. HFBRI-MAE~\cite{yin2025hfbri} combines handcrafted features with rotation-invariant representations but relies on traditional feature engineering. Most approaches treat masking as simple extensions of 2D methods, overlooking the potential of structured masking for rotation invariance. Recent advances in 2D vision like evolved part masking (EPM)~\cite{feng2023evolved} and attention-guided approaches like AttnMask~\cite{kakogeorgiou2022attnmask} suggest that structured masking can improve representation learning. However, the integration of structured spatial patterns with semantic-aware masking in rotation-invariant frameworks remains largely unexplored. 


\subsection{Rotation-Invariant Point Cloud Analysis}
In recent years, deep learning for 3D point clouds has matured substantially, with strong performance across classification, segmentation, detection, and registration~\cite{guo2020deep, ZhangCVPR21_SGGpoint, qi2017pointnet++, XiangICCV21CurveNet}.
However, robustness to arbitrary rotations remains a persistent bottleneck for real-world deployment, which motivates explicit rotation-invariant design.
Achieving rotation invariance is crucial for practical 3D applications where objects appear in arbitrary orientations. Handcrafted rotation-invariant features leverage geometric properties that remain unchanged under rotations. RIConv~\cite{zhang2019rotation} encodes local geometric relationships using distance and angle features. RIConv++~\cite{zhang2022riconv++} enhances this approach with global features. PaRot~\cite{zhang2023parot} disentangles pose information through patch-wise networks. RCPC~\cite{rcpc2024} addresses ambiguities via neighbor point sorting. ERINet~\cite{gu2021erinet} and its enhanced version~\cite{gu2022enhanced} combine local and global learning strategies for improved rotation-invariant representations. Orthogonal to rotation invariance, robustness to noise corruption has also been pursued; a spatial sorting and set-mixing aggregation module has been shown to enhance recognition under noisy perturbations~\cite{ZhangACCV24_SetMixer}.

Equivariant neural networks ensure theoretical rotation invariance through architectural design. Tensor Field Networks~\cite{thomas2018tensor} leverage spherical harmonics for SE(3)-equivariant representations. Alternative approaches include PCA-based canonical alignment~\cite{kim2020rotation,li2021closer} and LGR-Net~\cite{zhao2022rotation} that combines local-global features. Complementary to equivariant architectures and canonicalization, recent work rethinks rotation invariance through a registration-centric, staged framework \cite{YuAAAI23_RethinkRI}. While these strategies offer principled guarantees, they can introduce architectural complexity or alignment sensitivity in practice. Most methods face trade-offs between computational efficiency and robustness. Rather than modifying network architectures, we address this challenge at the masking strategy level, enabling plug-and-play integration with existing frameworks while achieving superior efficiency and broader applicability.

\section{Method}
\begin{figure*}[t]
    \centering
    \includegraphics[width=\textwidth]{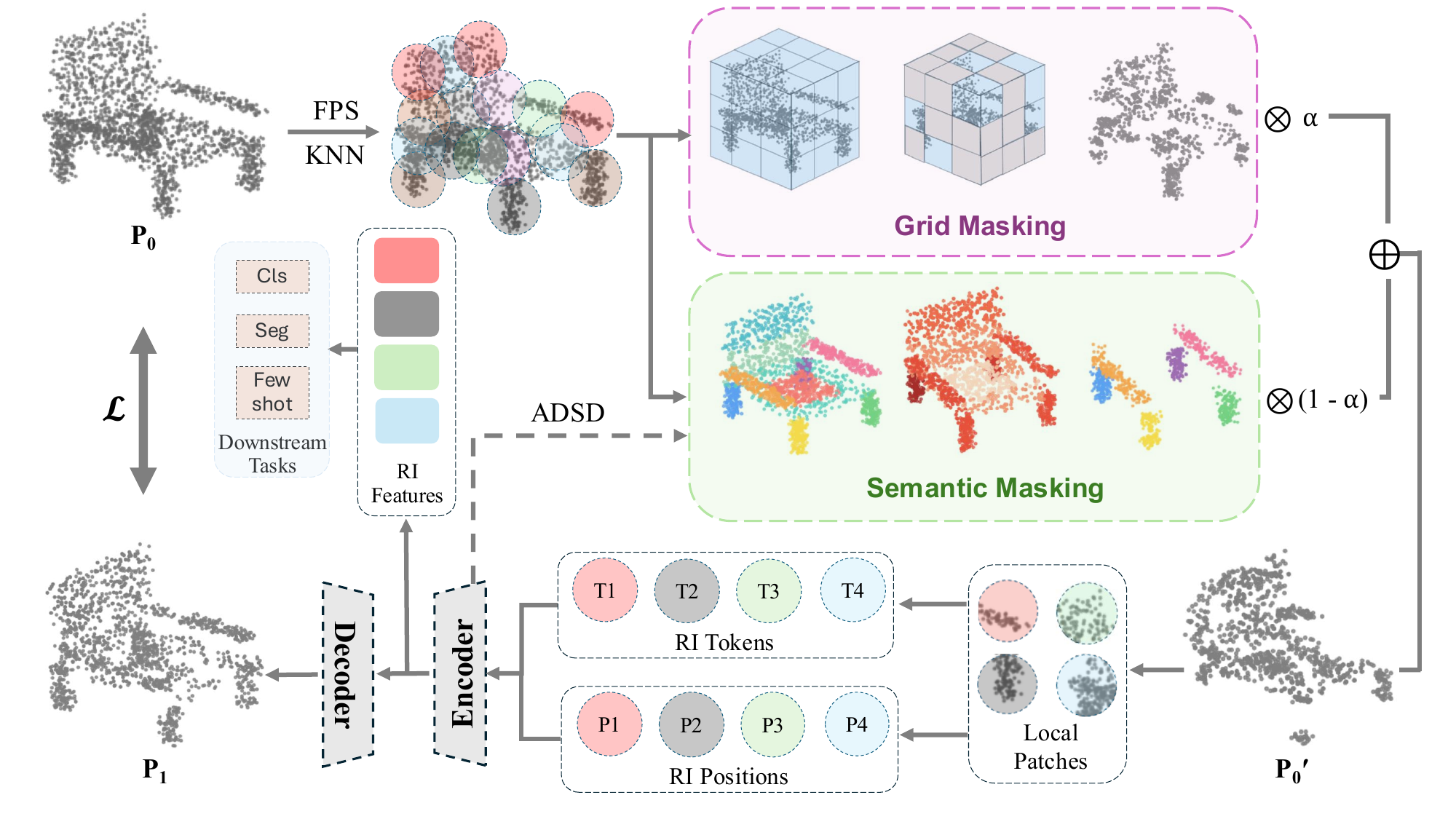}
    \caption{Overview of our dual-stream masking framework for rotation-invariant point cloud masked autoencoders. The input point cloud $P_0$ is processed through FPS and KNN to generate local patches, which are then processed by two complementary masking streams: Grid Masking creates structured spatial patterns based on coordinate organization, while Semantic Masking discovers coherent object parts through attention-driven clustering. These streams are dynamically weighted ($\alpha$ and $1-\alpha$) and combined to produce the final unmasked point cloud $P_0'$. The masked patches are processed through rotation-invariant (RI) feature extraction to generate RI tokens and RI positions, then fed into the RI MAE architecture for reconstruction learning. The decoder outputs the reconstructed point cloud $P_1$, and the extracted RI features can be utilized for various downstream tasks such as classification (Cls), segmentation (Seg), and few-shot learning.}
    \label{fig:structure}
\end{figure*}

\subsection{Problem Formulation}

Given a 3D point cloud $\mathbf{P} \in \mathbb{R}^{N \times 3}$ containing $N$ points in arbitrary orientation, our objective is to enhance rotation-invariant masked autoencoder frameworks through improved masking strategies. We define rotation invariance as the property where feature encoder $F$ produces consistent outputs:
\begin{equation}
F(\mathbf{P}) \equiv F(R\mathbf{P}), \quad \forall R \in SO(3),
\end{equation}
where $R$ represents any rotation matrix in the special orthogonal group $SO(3)$.

Current rotation-invariant MAE methods face a critical limitation in their masking strategies. Random masking fails to capture the geometric structure and semantic coherence essential for robust rotation-invariant learning. Our approach addresses this limitation through a novel dual-stream masking paradigm. This paradigm can be integrated into existing rotation-invariant MAE frameworks to enhance their performance.

\noindent\textbf{Dual-Stream Masking Strategy.}
We fuse two complementary streams to obtain per-patch masking probabilities:
\begin{equation}
\label{eq:mix}
\mathbf{M}^{(t)} \;=\; \big(1-\alpha(t)\big)\,\mathbf{M}_{\text{spatial}}
\;+\; \alpha(t)\,\mathbf{M}^{(t)}_{\text{semantic}},
\end{equation}
where $\mathbf{M}_{\text{spatial}}$ denotes the 3D Spatial Grid Masking that encodes geometric structure, and $\mathbf{M}^{(t)}_{\text{semantic}}$ is the Progressive Semantic Masking at training iteration $t$ that discovers semantic parts.
The curriculum coefficient $\alpha(t)\in[0,1]$ increases from geometric to semantic emphasis over training and is defined in Sec.~III-D.
This re-parameterization yields a normalized convex combination and avoids tuning two correlated weights.

This formulation creates a curriculum learning progression. Early training emphasizes structured geometric patterns through spatial grid masking. Later stages focus on semantic coherence through attention-driven discovery. 

The input point cloud $\mathbf{P}_0$ undergoes standard patch decomposition into $K$ local regions $\{\mathbf{P}_i\}_{i=1}^K$ using Farthest Point Sampling (FPS) for representative centroid selection. We then apply $k$-Nearest Neighbors (KNN) for coherent neighborhood formation.

Our masking strategies are designed as plug-and-play components. They enhance existing rotation-invariant MAE frameworks without requiring architectural modifications. This ensures broad applicability across different rotation-invariant approaches. As illustrated in Figure~\ref{fig:structure}, our dual-stream masking produces the unmasked point cloud $\mathbf{P}_0'$, where the unmasked patches are processed through rotation-invariant (RI) feature extraction to generate RI tokens and RI positions, which are then fed into the encoder to produce rotation-invariant features. These extracted features can be effectively utilized for various downstream tasks including classification, segmentation and few-shot learning. The decoder reconstructs the masked regions using the encoded features to produce the predicted point cloud $\mathbf{P}_1$, enabling the model to learn meaningful rotation-invariant representations through the reconstruction objective.

\subsection{3D Spatial Grid Masking}

Random masking in existing methods treats point cloud patches independently without considering their spatial relationships. This approach overlooks the geometric structure that remains consistent under rotations. Our 3D Spatial Grid Masking leverages these spatial relationships by creating structured masking patterns that preserve geometric consistency across orientations.

Rotation-invariant learning requires understanding spatial organization at both local and global levels. Local neighborhoods maintain their internal structure while their global arrangement follows predictable patterns. We exploit this property by organizing patches into spatial grids that reflect the underlying coordinate structure. This encourages the model to learn geometric relationships that are invariant to rotation.

\textbf{Spatial Organization and Ranking.}
We begin by extracting $K$ patch centers $\{c_i\}_{i=1}^K$ from point cloud $\mathbf{P} \in \mathbb{R}^{N \times 3}$ using Farthest Point Sampling (FPS)~\cite{qi2017pointnet++}. Each patch contains $k$ points selected via $k$-nearest neighbors around its center (typically $k = 32$). This creates coherent local regions from the irregular point distribution.

We establish spatial relationships between patches through coordinate-based ranking. Each patch center $c_i$ receives a position rank in each dimension:
\begin{equation}
\text{pos}_d[i] = \text{rank}(c_i^d), \quad d \in \{x, y, z\},
\end{equation}
where $c_i^d$ is the $d$-th coordinate of center $c_i$. This ranking transforms the irregular spatial distribution into ordered sequences. We then map these sequences to regular grid coordinates, enabling structured masking patterns on irregular point cloud data.

\textbf{Grid Construction and Probabilistic Selection.}
We construct a 3D grid system by dividing the ranked patch sequences into regular intervals. Each patch receives grid coordinates based on its spatial ranking:
\begin{equation}
\text{grid}_d[i] = \left\lfloor \frac{\text{pos}_d[i]}{G_d} \right\rfloor \bmod 2, \quad d \in \{x, y, z\},
\end{equation}
where $G_d$ controls the grid granularity in dimension $d$. The floor division groups consecutive patches into the same grid cell. The modulo operation creates binary coordinates that alternate between 0 and 1.

Each patch belongs to one of eight grid types based on its binary coordinates $(\text{grid}_x[i], \text{grid}_y[i], \text{grid}_z[i])$. These correspond to the eight vertices of a unit cube. We assign each grid type $t \in \{0, 1, \ldots, 7\}$ a masking probability $p_t \in [0, 1]$. The grid type index for patch $i$ is:
\begin{equation}
\text{grid\_type}[i] = \text{grid}_x[i] + 2 \cdot \text{grid}_y[i] + 4 \cdot \text{grid}_z[i].
\end{equation}

We select patches for masking based on their grid type probabilities and the target masking ratio. Patches with higher grid type probabilities are prioritized for masking. This ensures that the final masking pattern follows the desired spatial structure while maintaining the specified masking ratio.

This approach creates structured yet flexible masking patterns. The grid structure ensures spatial consistency while the probability-based selection introduces controlled variation. 
This promotes learning of geometric relationships that remain stable under rotation.

\subsection{Progressive Semantic Masking}

3D spatial grid masking provides structured geometric patterns but lacks semantic awareness of object parts and functional components. Our Progressive Semantic Masking discovers and masks semantically meaningful regions. This approach leverages orientation-agnostic part-based representations to improve rotation invariance.

\textbf{Attention-Driven Semantic Discovery (ADSD).} We extract attention maps $\mathbf{A}^{(t)} \in \mathbb{R}^{K \times K}$ from the final transformer encoder block at training iteration $t$. Element $A^{(t)}_{i,j}$ represents the attention weight between patches $i$ and $j$. The final encoder block provides the highest-level semantic understanding developed throughout the network hierarchy.

Attention mechanisms naturally capture semantic relationships between patches. Patches from the same semantic component exhibit stronger mutual attention weights. We construct semantic affinity graphs using adaptive thresholding:
\begin{equation}
w^{(t)}_{i,j} = \begin{cases}
A^{(t)}_{i,j} & \text{if } A^{(t)}_{i,j} > \tau^{(t)} \\
0 & \text{otherwise}.
\end{cases}
\end{equation}
The adaptive threshold $\tau^{(t)}$ increases progressively with training iterations. This preserves only the strongest semantic connections as the model's understanding matures.

\textbf{Progressive Component Clustering.} We partition patches into $C^{(t)}$ semantic components using Expectation-Maximization (EM) algorithm based on the semantic affinity graph. The number of components decreases linearly with training progress:
\begin{equation}
C^{(t)} \;=\; C_{\max} - \frac{t}{T}\,\big(C_{\max}-C_{\min}\big).
\end{equation}
We set $C_{\text{max}} = 40$ and $C_{\text{min}} = 10$ based on empirical validation. This progressive reduction forces the clustering algorithm to identify increasingly coherent semantic parts. The process transitions from fine-grained local features to coarse-grained object-level components.

Each patch follows a multivariate Gaussian distribution within its component. The EM algorithm updates iteratively:

\textbf{E-step:} $\gamma^{(t)}_{i,c} = \frac{\pi^{(t)}_c \mathcal{N}(v_i|\mu^{(t)}_c, \Sigma^{(t)}_c)}{\sum_{j=1}^{C^{(t)}} \pi^{(t)}_j \mathcal{N}(v_i|\mu^{(t)}_j, \Sigma^{(t)}_j)}$.

\textbf{M-step:} Update parameters $\pi^{(t)}_c$, $\mu^{(t)}_c$, $\Sigma^{(t)}_c$ based on $\gamma^{(t)}_{i,c}$.

The attention feature vector $v_i = A^{(t)}_{i,:}$ captures patch $i$'s semantic relationships with all other patches. Component assignment follows: $C^{(t)}_i = \arg\max_k \gamma^{(t)}_{i,k}$.

\textbf{Semantic Coherence Masking.} Patches within the same semantic component receive identical masking probabilities:
\begin{equation}
\mathbf{M}^{(t)}_{\text{semantic}}[i] = \delta_{C^{(t)}_i}.
\end{equation}
Here $\delta_{C^{(t)}_i}$ is a random value uniformly sampled from $[0,1]$ and assigned to component $C^{(t)}_i$. Semantically related patches are masked or preserved together. This forces the model to learn complete semantic parts rather than fragmented regions. The strategy promotes part-based understanding that remains invariant under rotation since semantic object parts maintain their identity regardless of orientation.

\begin{table*}[htbp]
\centering
\caption{Classification accuracy (\%) of SVM classifiers trained on features extracted from ShapeNet-pretrained models, evaluated across different rotation settings on ModelNet40, ScanObjectNN, and OmniObject3D datasets.}
\label{tab:classification_results}
\small
\resizebox{\textwidth}{!}{ 
\begin{tabular}{lccccccccccccccc}
\toprule
\multirow{2}{*}{Method} & \multicolumn{5}{c}{ModelNet40}  & \multicolumn{5}{c}{ScanObjectNN} & \multicolumn{5}{c}{OmniObject3D}\\
\cmidrule(lr){2-6}
\cmidrule(lr){7-11} 
\cmidrule(lr){12-16} 

& A/A & A/R & Z/Z & Z/R & R/R & A/A & A/R & Z/Z & Z/R & R/R & A/A & A/R & Z/Z & Z/R & R/R\\
\midrule
RI-MAE~\cite{su2025ri} & 87.9 & 88.2 & 88.2 & 88.0 & 88.5 & 90.1 & 89.9 & 90.2 & 90.3 & 90.1& 73.8 & 73.0 & 73.3 & 73.5 & 73.2\\
With Ours & \cellcolor{gray!15}88.3 & \cellcolor{gray!15}88.5 & \cellcolor{gray!15}88.7 & \cellcolor{gray!15}88.5 & \cellcolor{gray!15}88.6 & \cellcolor{gray!15}90.5 & \cellcolor{gray!15}90.5 & \cellcolor{gray!15}90.2 & \cellcolor{gray!15}90.3 & \cellcolor{gray!15}90.3 & \cellcolor{gray!15}75.8 & \cellcolor{gray!15}75.2 & \cellcolor{gray!15}74.9 & \cellcolor{gray!15}75.3 & \cellcolor{gray!15}75.0\\
\midrule
MaskLRF~\cite{feng2023masklrf} & 88.3 & 88.5 & 88.2 & 88.7 & 88.7 & 89.9 & 90.2 & 90.2 & 90.1 & 89.8 & 75.9 & 74.9 & 75.1 & 75.7 & 75.5\\
With Ours & \cellcolor{gray!15}88.7 & \cellcolor{gray!15}89.1 & \cellcolor{gray!15}89.3 & \cellcolor{gray!15}89.1 & \cellcolor{gray!15}88.9 & \cellcolor{gray!15}90.1 & \cellcolor{gray!15}90.3 & \cellcolor{gray!15}90.4 & \cellcolor{gray!15}90.1 & \cellcolor{gray!15}90.1 & \cellcolor{gray!15}76.9 & \cellcolor{gray!15}76.2 & \cellcolor{gray!15}76.4 & \cellcolor{gray!15}76.3 & \cellcolor{gray!15}76.3\\
\midrule
HFBRI-MAE~\cite{yin2025hfbri} & 89.7 & 89.8 & 89.9 & 89.6 & 89.5 & 90.0 & 90.3 & 90.3 & 90.1 &  90.1 & 73.1 & 73.1 & 72.5 & 72.7 & 73.1\\
With Ours & \cellcolor{gray!15}89.9 & \cellcolor{gray!15}90.0 & \cellcolor{gray!15}90.1 & \cellcolor{gray!15}90.0 & \cellcolor{gray!15}89.8 & \cellcolor{gray!15}90.6 & \cellcolor{gray!15}90.7 & \cellcolor{gray!15}90.5 & \cellcolor{gray!15}90.2 & \cellcolor{gray!15}90.5 & \cellcolor{gray!15}75.0 & \cellcolor{gray!15}74.2 & \cellcolor{gray!15}74.9 & \cellcolor{gray!15}74.7 & \cellcolor{gray!15}74.7\\
\bottomrule
\end{tabular}
}
\normalsize
\end{table*}

\subsection{Dynamic Weighting Mechanism}

The combination of 3D spatial grid masking and progressive semantic masking requires a principled approach to balance their contributions throughout training. We introduce a dynamic weighting mechanism that adaptively transitions from geometry-focused learning to semantics-focused learning, implementing a curriculum learning~\cite{bengio2009curriculum} approach with three distinct phases.

The final masking probability for patch $i$ at training iteration $t$ is computed as a weighted combination of both masking strategies:
\begin{equation}
P^{(t)}_i = (1 - \alpha^{(t)}) \times P^{\text{grid}}_i + \alpha^{(t)} \times P^{(t)}_{\text{sem},i},
\end{equation}
where $\alpha^{(t)}$ is a time-dependent weighting parameter that follows a progressive schedule:
\begin{equation}
\alpha^{(t)} = \left(\frac{t}{T}\right)^{\gamma},
\end{equation}
where $T$ is the total number of training iterations and $\gamma$ controls the transition curve steepness. 

The dynamic weighting creates a curriculum learning~\cite{bengio2009curriculum} approach with three distinct phases. During the early stage ($\alpha^{(t)} \approx 0$), spatial grid masking dominates to learn fundamental geometric relationships and low-level spatial features. The transition stage ($\alpha^{(t)} \approx 0.5$) balances both strategies, bridging geometric understanding with emerging semantic awareness. In the late stage ($\alpha^{(t)} \approx 1$), progressive semantic masking takes over to focus on high-level semantic relationships and part-based understanding.

\section{Experiments and Results}
\begin{figure*}[htbp]
    \centering
    \includegraphics[width=\textwidth]{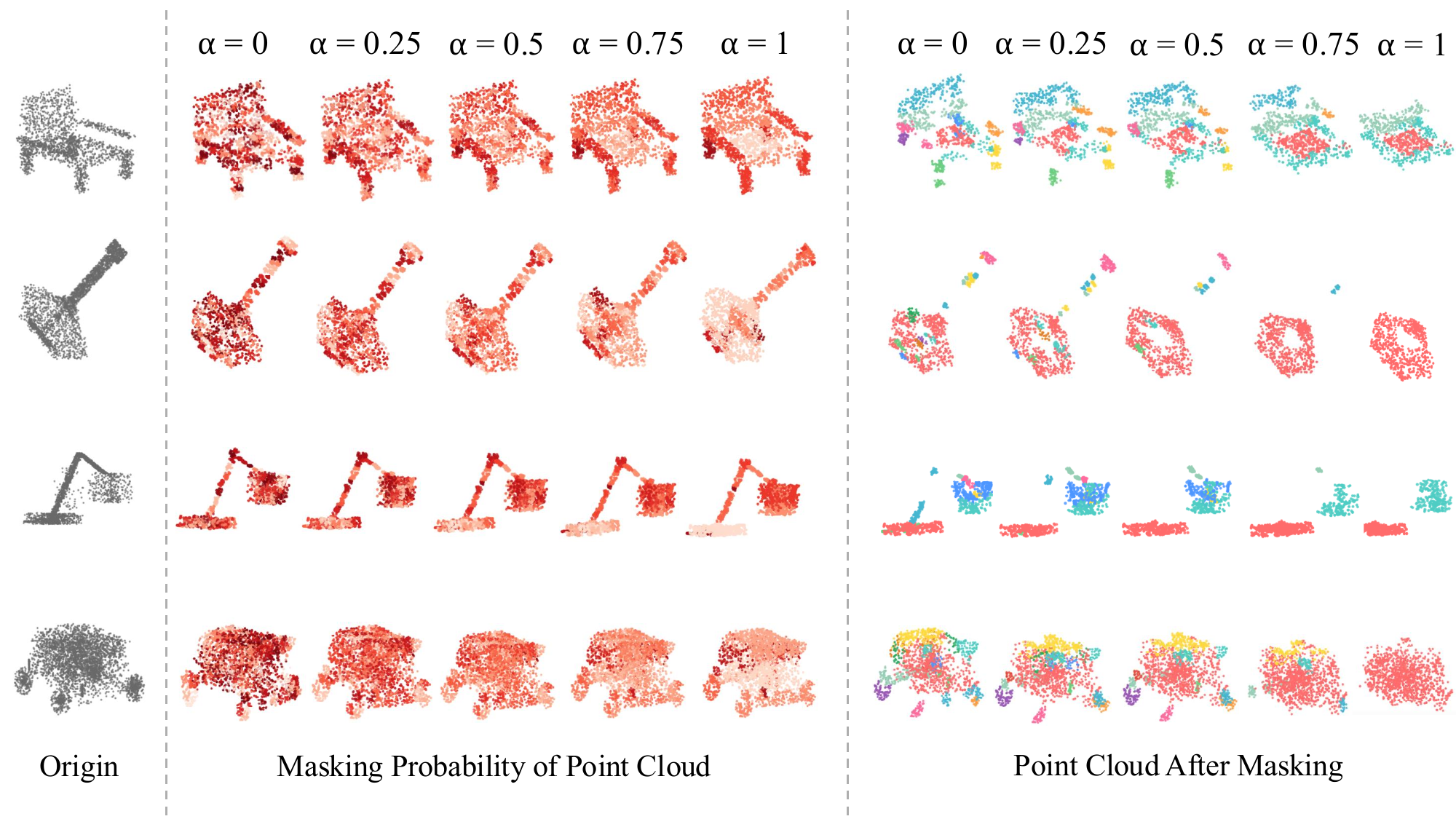}
    \caption{Visualization of masking patterns with varying $\alpha$ values. Left: Masking probability visualization showing the transition from grid-based ($\alpha=0$) to semantic-based ($\alpha=1$) patterns. Right: Actual masked point clouds demonstrating how different $\alpha$ values affect the final masking results across various object categories.}
    \label{fig:vis}
\end{figure*}

\subsection{Datasets}

We conduct comprehensive experiments using ShapeNet~\cite{chang2015shapenet} for pretraining and evaluate on three standard benchmarks: ModelNet40~\cite{wu20153d}, ScanObjectNN~\cite{uy2019revisiting}, and OmniObject3D~\cite{wu2023omniobject3d}. ShapeNet contains 51,300 models across 55 categories with 2,048 points per instance, providing rich geometric diversity for self-supervised pretraining. For evaluation, ModelNet40 comprises 40 categories with 1,024 points per model, serving as a standard classification benchmark. ScanObjectNN contains real-world scanned objects with background noise and occlusions, offering a more challenging evaluation scenario. Additionally, OmniObject3D provides a large-scale multi-modal 3D dataset with over 6,000 scanned objects across 190 categories, featuring high-quality meshes and realistic textures for comprehensive evaluation.

\subsection{Experimental Setup}

\textbf{Implementation Details.} We implement our dual-stream masking strategies and integrate them with existing rotation-invariant MAE frameworks using PyTorch, training on NVIDIA RTX 4090 GPUs. 

Our dual-stream masking strategy combines 3D spatial grid masking and progressive semantic masking with dynamic weighting parameter $\gamma=2$. Grid granularity parameters are set to $G_x = G_y = G_z = 4$. The semantic masking component employs EM clustering with component numbers linearly decreasing from $C_{\text{max}}=40$ to $C_{\text{min}}=10$. The overall masking ratio is maintained at 75\%.

We pretrain on ShapeNet for 300 epochs using AdamW optimizer with learning rate $lr=1.5 \times 10^{-4}$, weight decay of $1 \times 10^{-2}$, and cosine annealing scheduler. The batch size is set to 128 with mixed precision training.

\textbf{Baseline Methods.} We evaluate our masking strategies by integrating them with state-of-the-art rotation-invariant MAE frameworks: HFBRI-MAE~\cite{yin2025hfbri}, MaskLRF~\cite{feng2023masklrf} which employs local reference frame-based approaches, and RI-MAE~\cite{su2025ri} representing an alternative rotation-invariant design. This plug-and-play evaluation demonstrates the generalizability of our masking approach across different architectural frameworks.

\textbf{Evaluation Protocol.} To validate rotation invariance capabilities, we evaluate across three distinct rotation configurations: Aligned ($A$) maintains canonical upright positioning, Z-rotated ($Z$) applies random rotations around the z-axis, and Random ($R$) employs arbitrary 3D rotations in SO(3) space. We conduct experiments on five rotation scenarios where X/Y denotes pretraining/testing orientations: $A/A$, $A/R$, $Z/Z$, $Z/R$, and $R/R$.

We extract features from pretrained encoders and train linear SVM classifiers for downstream classification. Classification accuracy is reported across all rotation settings to demonstrate the effectiveness of our masking strategies in enhancing rotation-invariant representation learning.


\subsection{Experimental Results}

\begin{figure}[t]
    \centering
    \includegraphics[width=\columnwidth]{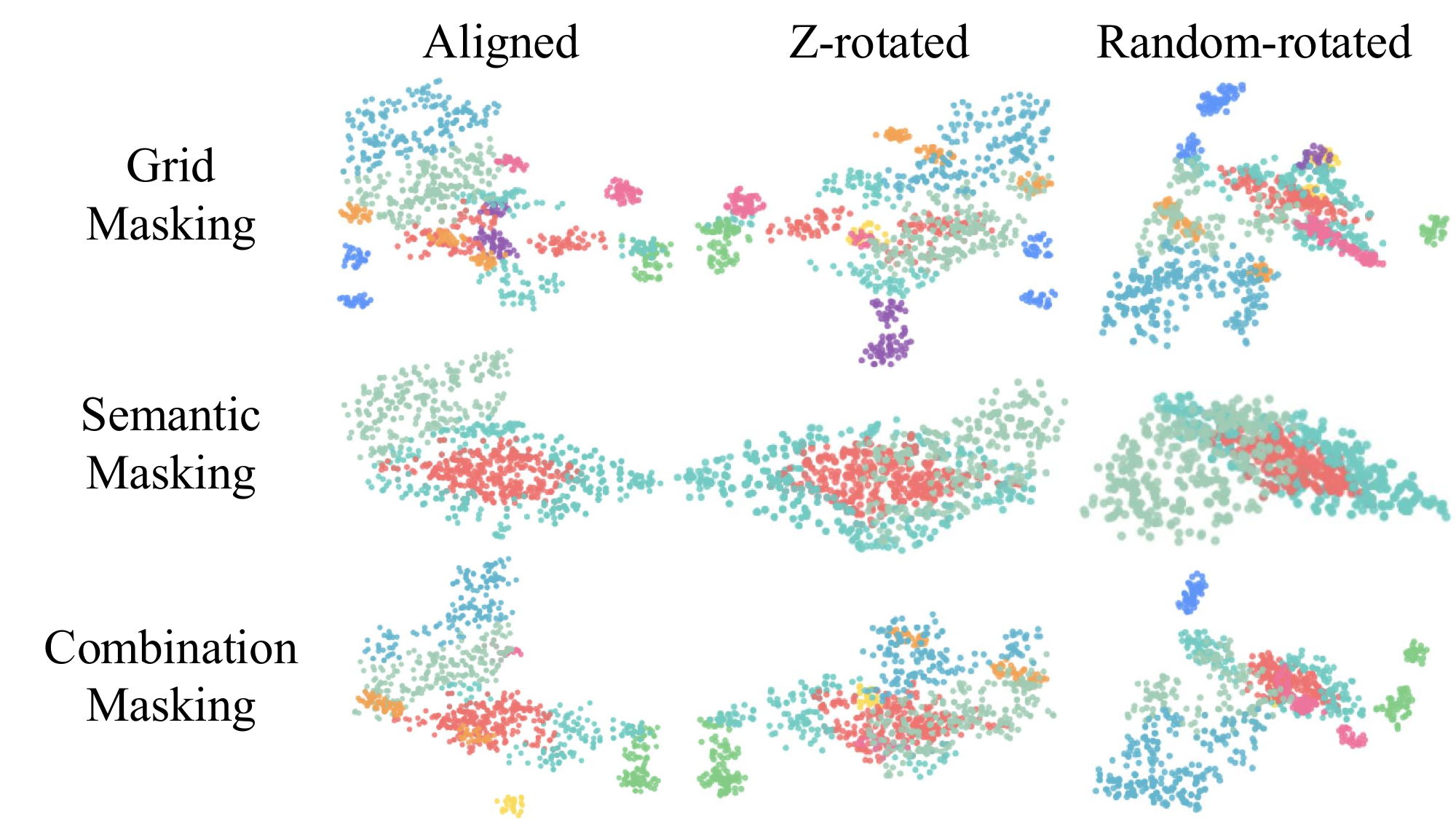}
    \caption{Qualitative comparison of masking strategies across rotation scenarios. Our approach maintains consistent masking quality for (top) Grid Masking based on spatial structure, (middle) Semantic Masking based on attention-driven clustering, and (bottom) Combination Masking across aligned, Z-rotated, and random-rotated point clouds.}
    \label{fig:rotation}
\end{figure}

Table~\ref{tab:classification_results} presents the classification accuracy when state-of-the-art rotation-invariant MAE frameworks are equipped with our dual-stream masking strategy across three datasets and five rotation scenarios. Our masking approach demonstrates consistent improvements over the baseline methods, particularly in challenging rotation-invariant scenarios.

In addition to the averaged gains in Table~I, we note a clear dependence on the pose distribution: improvements are amplified when rotations span broad $SO(3)$ ranges and attenuated under constrained or near-aligned settings. This pattern aligns with the intended roles of the two streams—the grid-based branch stabilizes long-range geometry across poses, while the semantics-informed branch preserves part coherence once attention becomes reliable. On cleaner, object-centric benchmarks the effect is more pronounced, whereas cluttered real scans yield smaller yet consistent gains; qualitative visualizations across rotation scenarios further corroborate this trend. Taken together, the masking prior steers pretext learning toward pose-stable signals without modifying the backbone, thus complementing standard random masking.

\subsection{Ablation Studies}

\textbf{Component Number Analysis.} We analyze the impact of fixed component numbers versus dynamic scheduling in Table~\ref{tab:ablation}. The results demonstrate that dynamic component scheduling with $\gamma=2$ achieves optimal performance across all datasets.

Beyond the aggregate outcome that dynamic scheduling with $\gamma{=}2$ attains the best average accuracy across datasets, the progression from finer early partitions to later consolidation appears crucial for sample-efficient part discovery. Smaller $\gamma$ values tend to shift emphasis toward semantics prematurely, while larger $\gamma$ delays consolidation and underutilizes reliable attention cues; the transient increase in component count adds only mild clustering overhead and leaves the backbone unchanged. Practically, a moderately convex schedule thus provides a stable handoff from geometric regularities to semantic structure, consistent with the curriculum implied by our attention-driven masking.

\begin{table}[htbp]
\centering
\caption{Average accuracy across 5 rotation settings on component number settings and dynamic weighting parameter $\gamma$ based on HFBRI-MAE.}
\label{tab:ablation}
\small
\resizebox{\columnwidth}{!}{%
\begin{tabular}{ccccc}
\toprule
$C$ & $\gamma$ & ModelNet40 & ScanObjectNN &  OmniObject3D\\
\midrule
5   & 2   & 89.6 & 89.7 & 73.9 \\
10 & 2   & 89.8 & 89.9 & 74.3 \\
20  & 2   & 89.4 & 89.5 & 73.9 \\
\midrule
\multirow{4}{*}{dynamic (5→10)} 
& 0.5 & 89.7 & 89.9 & 73.8 \\
& 1   & 89.9 & 90.2 & 74.5 \\
& 2   & \cellcolor{gray!15}90.0& \cellcolor{gray!15}90.4& \cellcolor{gray!15}74.6 \\
& 5   & 89.8 & 90.0 & 74.4 \\
\bottomrule
\end{tabular}
}
\end{table}

\subsection{Qualitative Analysis}

Figure~\ref{fig:vis} illustrates the effect of different $\alpha$ values on masking patterns. As $\alpha$ increases from 0 to 1, the masking strategy transitions from purely grid-based (left) to purely semantic-based (right), demonstrating the curriculum learning progression from geometric structure understanding to semantic part discovery.

Figure~\ref{fig:rotation} provides a qualitative comparison of our masking strategies across different rotation scenarios, demonstrating the robustness of our approach to orientation changes.

\subsection{Training Time and Inference Cost}
Our dual-stream masking introduces lightweight grid ranking and attention-guided clustering during pretraining, which yields a modest wall-clock increase of 12.8--15.0\% across RI-MAE, MaskLRF, and HFBRI-MAE (Table~\ref{tab:time}). 
Crucially, the encoder/decoder architectures remain unchanged; the masking strategy only governs which patches are visible during training. 
Therefore, inference cost is \emph{identical} to each backbone at test time, since downstream evaluation extracts features with the same encoders without any masking or extra modules. 
In practice, the added cost stems from per-batch spatial ranking and a small-component EM over patch-level attentions with a decreasing schedule, both implemented outside the forward path. 
This aligns with our plug-and-play design that improves rotation robustness while preserving inference efficiency.

\begin{table}[htbp]
\centering
\caption{Wall-clock training time (hh:mm) under Random Masking vs.\ our dual-stream masking across three RI backbones; $\Delta$ is the percentage increase. Inference time remains unchanged.}
\label{tab:time}
\small
\begin{tabular}{lccc}
\toprule
Training Time & RI-MAE & MaskLRF & HFBRI-MAE  \\
\midrule
Random Masking & 21h 36min & 25h 08min & 22h 49min  \\
With Ours      & 24h 51min & 28h 21min & 26h 01min  \\
$\Delta$       & \(+15.0\%\) & \(+12.8\%\) & \(+14.0\%\) \\
\bottomrule
\end{tabular}
\normalsize
\end{table}
\section{Conclusion}
We presented a plug-and-play dual-stream masking paradigm for rotation-invariant point cloud MAE, combining coordinate-based 3D Spatial Grid Masking with attention-driven Progressive Semantic Masking under curriculum learning. The design injects a targeted spatial--semantic bias while remaining architecture-agnostic. Experiments show consistent, albeit modest, improvements across datasets and rotation scenarios. Complementary to random masking, our approach will be further validated via multi-source pretraining and noise-robust variants to better handle real-world scans while quantifying cost--performance trade-offs. 
Additionally, we will explore cross-domain self-distillation to broaden generality and investigate lightweight clustering/ranking to keep masking overhead small at scale. We view this as a simple and broadly applicable plugin toward robust rotation-invariant point cloud pretraining.

{\small
\bibliographystyle{IEEEtran}
\bibliography{IEEEexample}
}


\end{document}